\documentclass{article} 
\usepackage{aperture_arxiv,times}

\usepackage{amsmath,amsfonts,bm}

\def\eqref#1{equation~\ref{#1}}

\def\1{\bm{1}}

\DeclareMathAlphabet{\mathsfit}{\encodingdefault}{\sfdefault}{m}{sl}
\SetMathAlphabet{\mathsfit}{bold}{\encodingdefault}{\sfdefault}{bx}{n}

\usepackage{etoolbox}
\usepackage{hyperref}
\usepackage{url}
\usepackage{booktabs}
\usepackage{graphicx}
\usepackage{colortbl}  
\usepackage{pifont}     
\usepackage{multirow}
\usepackage{float}
\usepackage{needspace,ragged2e}
\usepackage{wrapfig,needspace,ragged2e}
\usepackage{tabularx}
\usepackage{wrapfig}
\usepackage{caption}
\usepackage{hyperref}
\usepackage{cleveref}
\usepackage{tikz}
\usetikzlibrary{tikzmark}
\definecolor{ourblue}{RGB}{20,60,170}
\definecolor{deltared}{RGB}{190,20,30}
\newtoggle{CommentsMode}
\toggletrue{CommentsMode}
\togglefalse{CommentsMode}
\definecolor{deepgreen}{RGB}{34,139,34}
\definecolor{nbpurple}{RGB}{104,72,143}
\definecolor{nborange}{RGB}{178,92,24}

\title{Aperture: Training-Free Multiscale Concept Bottlenecks for Remote Sensing}

\definecolor{apertureprojectred}{HTML}{8B1E2D}

\author{%
  \parbox{\dimexpr\textwidth-2\tabcolsep\relax}{\centering
    \normalfont\fontsize{9}{11}\selectfont
    {\hypersetup{urlcolor=black}%
    \href{https://rishabh-mondal.github.io/}{\textbf{Rishabh Mondal}}\textsuperscript{1,2}\quad
    \href{https://nipunbatra.github.io/}{\textbf{Nipun Batra}}\textsuperscript{1}\quad
    \href{https://utkarshmall.com/}{\textbf{Utkarsh Mall}}\textsuperscript{2}\\[5pt]}
    \fontsize{8}{9.5}\selectfont
    \textsuperscript{1}Indian Institute of Technology Gandhinagar, India\\
    \textsuperscript{2}Mohamed bin Zayed University of Artificial Intelligence, UAE\\[10pt]
    {\fontsize{9}{11}\selectfont\bfseries
    \href{https://rishabh-mondal.github.io/aperture/}{%
      \textcolor{apertureprojectred}{Website: APERTURE}}}
  }%
}

\iclrfinalcopy
\begin{document}

\maketitle
\newcommand{\methodname}{\textsc{Aperture}}
\begin{abstract}
While earth observation models have advanced substantially, they still lack interpretability.
While concept-bottleneck models provide interpretability and expert interaction, they are either too expensive to train for the remote sensing domain or perform poorly without annotation.
We posit that in expert domains like remote sensing, such training-free models require both fine details in both image and concept space.
In image space, we propose a multiscale concept bottleneck using greedy quadtree routing to locate small concepts.
In concept space, we replace contrastive vision language models with pre-trained MLLMs and present a way to get reliable concept scores from them.
We introduce \methodname~that blends concept scores at the global image and native concept-scale level to give state-of-the-art training-free model performance.
To test these models, introduce SiFC, a fine-grained concept-centric dataset across three countries, with human-reviewed class-level concept maps. 
On SiFC, \methodname ~outperforms the best training-free baselines by more than 10 percentage points in macro F1-score, and notably also outperforms supervised concept bottleneck models.
Targeted component-removal tests examine whether concept scores respond to changes in visual evidence, while temporal experiments show that descriptor updates improve recognition of technological changes without retraining.
\end{abstract}

\section{Introduction}
\label{sec:introduction}

\begin{figure}[h]
    \centering

    \begin{minipage}[t]{0.6\linewidth}
        \vspace{0pt}
        \centering
        \includegraphics[width=\linewidth]{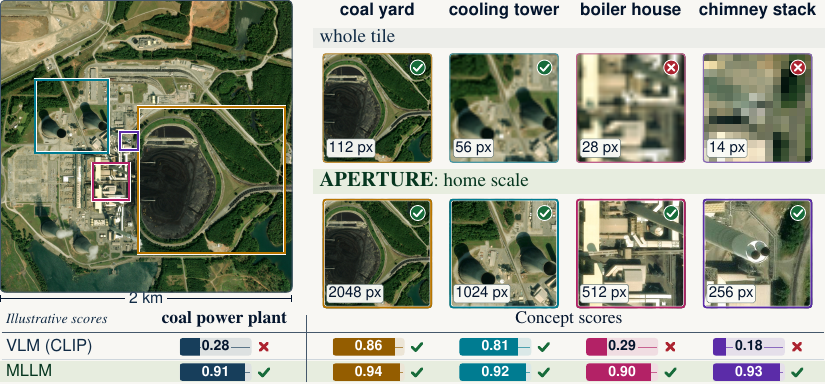}
    \end{minipage}
    \begin{minipage}[t]{0.32\linewidth}
        \vspace{0pt}
        \setlength{\abovecaptionskip}{0pt}
        \setlength{\belowcaptionskip}{0pt}
        \caption{\textbf{Multi-scale concept bottleneck:} Resizing the $4096$-pixel coal plant (spanning $4\mathrm{km}^2$) shrinks the coal yard and chimney to $112$ and $14$ pixels (top right). 
        \textsc{Aperture} instead crops concepts at home scales, enabling it to recognize both parts and correctly classify the site, unlike CLIP.}
        \label{fig:aperture-teaser}
    \end{minipage}

\end{figure}




Earth observation enables continuous monitoring  of various events and concepts, such as deforestation or the construction of new industries, so that rapid actions can be taken to protect our environment.
Visual recognition in satellite imagery has advanced substantially over the last decade, driven by progress in fine-grained~\citep{chu2024fine}, self-supervised~\citep{ayush2021geography,cong2022satmae}, vision-language~\citep{li2023rs,hu2025rsgpt}, and reasoning models~\citep{yao2026remotereasoner,liu2026towards}.
Yet, interpretable models remain underexplored. 
Remote sensing is an expert domain that requires close interaction between model users (domain experts) and models.
Interpretable models, especially ante-hoc (or interpretable-by-design) models, are specifically useful in such cases, as the experts can not only understand the model's decision but can also intervene and improve them.

Concept bottleneck models (CBMs)~\citep{pmlr-v119-koh20a} are one way to create such ante-hoc models. 
CBMs make class recognition decisions on top of an intermediate list of human-understandable concepts, making the decision more transparent and intervenable.
However, they require not only class-level but also expensive concept-level annotations, making data collection significantly time-consuming~\citep{Mall_2021_ICCV}.
As a result, several follow-up works have proposed both concept label-free~\citep{oikarinen2023labelfree} and completely training-free variants~\cite{menon2023visual}.

Training-free CBMs are ideal for expert domains such as ecology or remote sensing, as often we do not have extensive labeled training data, and novel classes can arise at deployment time.
They allow for easy class and concept addition at deployment and are therefore ideal for this scenario.
However, existing training-free CBM methods still struggle when applied to satellite image recognition.

Firstly, training-free CBMs rely on contrastive vision-language models such as CLIP~\citep{radford2021learning} to recognize concepts. 
But these models struggle with non-object-centric attribute recognition~\citep{kang2025clip}, especially in fine-grained expert domains such as remote sensing.
We address this issue by replacing contrastive VLMs with a multimodal large-language model (MLLM)-based formulation.
This enables us to reason better about domain-specific concepts and attributes.

Using an MLLM inside a CBM introduces a second, non-trivial problem.
Generative MLLMs naturally return discrete presence/absence decisions, whereas CBMs rely on continuous concept scores.
Asking an MLLM to further explicitly report its confidence results in unreliable and miscalibrated scores.
Borrowing from binary question answering works~\citep{giovannotti2024calibrated}, we derive continuous concept evidence from the normalized next-token probabilities assigned to \texttt{Yes} and \texttt{No}.
This yields significantly more effective concept scores.
To our knowledge, we are the first to apply MLLMs to training-free CBMs and use the decoding probability to score concept confidence.

Satellite imagery introduces a third challenge: relevant concepts may occupy vastly different spatial extents.
For example, as shown in \cref{fig:aperture-teaser}, coal power plant may span $4 km^2$, but an important concept, such as a chimney stack, only occupies $0.016  km^2$ ($<0.4\%$ of the image).
This makes recognition of these small concepts very difficult.
Therefore, we introduce a multi-scale concept recognition framework.
Instead of just looking for a concept at the global image scale, we recursively break down images into smaller scales to search and aggregate concept scores across them.
However, exhaustively evaluating every crop is computationally prohibitive.
Recursively splitting an image into 4 smaller quadrants 4 times (see ~\cref{fig:aperture-teaser}), requires evaluating 341$\times$ more regions than global evaluation.
Therefore, we introduce a greedy quadtree search that recursively searches within only one crop at every zoom level, reducing the search time from exponential to linear in quadtree depth.


More importantly, satellite imagery provides information typically unavailable in ordinary internet imagery: \emph{known ground sampling distance (GSD)}.
We exploit this information to associate each concept with a characteristic physical ``home scale'', allowing the search to stop at the resolution at which the concept should be naturally recognized.

%
%
%

These limitations are difficult to study with the existing remote-sensing benchmarks, which mostly contain atomic scene categories~\citep{8736785, sumbul2019bigearthnet},
recognizable without identifying constituent parts and attributes.
We therefore introduce a new dataset, SiFC (Satellite image Facility Concepts), with 800 images from 6 fine-grained industrial classes.
We additionally construct a subset of FMoW~\citep{christie2018functional}, consisting of similar multi-concept fine-grained categories.
Both these datasets span multiple world regions to test out-of-region generalization.

Together, these components turn training-free CBMs from a global contrastive-similarity model into a generative, spatially adaptive, and physically grounded concept recognition framework for satellite imagery that we call \methodname.
We show that \methodname~ achieves state-of-the-art performance among prior training-free CBMs on both SiFC and FMoW.
Notably, \methodname~ even outperforms supervised CBMs on out-of-region images.
We also show how our training-free method can easily allow interventions from experts when a class concept changes over time.
Our contributions are:

\noindent$\bullet$ We introduce an MLLM-based training-free concept bottleneck for fine-grained satellite recognition and derive continuous concept evidence from binary next-token probabilities for it.

\noindent$\bullet$ We introduce efficient greedy quadtree routing for discovering concepts that may occupy only a small fraction of the full image.

\noindent$\bullet$ We introduce \emph{home scale}, exploiting known GSD in satellite images to associate concepts with their characteristic physical extent and constrain multiscale recognition to an appropriate resolution.

\noindent$\bullet$ We introduce a concept-centric benchmark, SiFC, where we achieve state-of-the-art training-free performance, even outperforming supervised interpretable models under geographic shifts

\section{Related Works}
\noindent\textbf{Interpretable recognition:} Post-hoc methods, such as Grad-CAM~\citep{selvaraju2017grad,8354201}, explain predictions without guaranteeing faithful insight into the model's decision process~\citep{adebayo2018sanity,rudin2019stop}. 
This motivated ante-hoc models that incorporate human-readable concepts into the prediction process.~\citep{chen2019looks}. 
Among ante-hoc methods, Concept Bottleneck Models (CBMs)~\citep{pmlr-v119-koh20a,espinosa2022concept,pmlr-v202-kim23g,espinosa2023learning} have become a prominent approach in expert domains, as they provide an intermediate human-interpretable concept layer.
However, standard CBMs require both class labels and image-level concept annotations, the latter of which are significantly expensive to collect at scale~\citep{Mall_2021_ICCV}. 
Label-free CBMs remove concept annotations through pretrained vision-language models but still require labeled images to learn the class predictor~\citep{oikarinen2023labelfree,yang2023language,srivastava2024vlg}. 
More recently, entirely training-free CBMs, such as classification-by-description~\citep{menon2023visual, pratt2023does, chen2025interpretable,yamaguchi2025zero}, have been introduced. 
They remove target-specific fitting entirely and require neither labeled images nor image-level concept annotations.
We show that existing training-free CBMs struggle with fine-grained interpretable recognition in satellite imagery and introduce a new framework for it.
While our work explores a spatial hierarchy over image regions, prior works have explored semantic hierarchies among concepts~\citep{pittino2023hierarchical,hill2026hierarchical}.


\noindent\textbf{Interpretable recognition in satellite imagery:} 
Concept-based explanations of satellite scene classification aim to link class decisions to human-interpretable visual concepts.
Existing methods highlight influential regions~\citep{kakogeorgiou2021evaluating} or generate textual reasoning before answering visual questions~\citep{earthvqa}.
However, these approaches do not explicitly expose how intermediate concepts contribute to the prediction.
PCINet~\citep{hua2024pcinet} is related to our setting, using LLM-generated concepts for interpretable aerial-image recognition.
However, it evaluates on MAI dataset~\citep{hua2021aerial}, a scene-recognition benchmark with broad categories.
In contrast, SiFC studies fine-grained compositional recognition, where competing system-level facility classes share constituent parts and attributes, and recognition therefore requires distinguishing classes through their combinations.
Furthermore, SiFC is constructed with the same classes represented across three geographic regions, enabling evaluation of out-of-region generalization.
\section{Satellite Imagery Facility Concepts (SiFC)}
\label{sec:sifc}

\begin{figure}[h]
    \centering
    \includegraphics{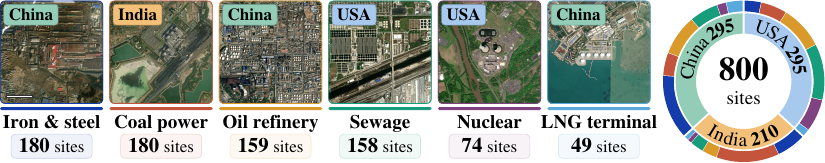}
    \caption{\textbf{SiFC at a glance.} (Left) sample facility classes, labeled with their country and the number of sites. Each tile covers about 4$km^2$ (4096$\times$4096 px at zoom level 18, World Imagery Wayback~\citep{esri_home} ); the white bar is 500\,m. (Right) 800 sites by country (inner) and classes (outer).}
    \label{fig:sifc-overview}
\end{figure}

Existing recognition datasets for remote sensing largely focus on very coarse scene or land-cover categories, which do require limited reasoning about parts and their attributes. For example, widely used datasets such as EuroSAT~\citep{8736785} and BigEarthNet~\citep{sumbul2019bigearthnet} primarily contain land-cover categories that are not naturally characterized through a composition of semantically meaningful concepts. 
Even higher-resolution scene-recognition datasets such as RESISC45~\citep{cheng2017remote} include many relatively atomic categories, such as baseball fields.
In contrast, we seek \emph{system}-like entities that can be characterized by the presence, arrangement, and attributes of their constituent components. 
Industrial facilities naturally fit this setting: steel and power plants contain recurring functional sub-components that support recognition by parts and attributes.

To support this setting, we introduce \textbf{Satellite Imagery Facility Concepts (SiFC)}, a dataset for concept-based classification of industrial facilities in satellite imagery. 
SiFC contains 800 samples from \emph{three regions}: USA, India, and China and covers \emph{six facility} classes. 
This geographic diversity captures variation in facility appearance, surrounding land use, and component layout
Images are sourced from the public ESRI World Imagery Wayback archive. 
We release corner coordinates with retrieval scripts rather than redistribute imagery.
Each image has a resolution of $4096 \times 4096$ pixels and is retrieved at zoom level 18, covering an area of approximately $4\,\mathrm{km^2}$.
Samples include facility and country labels, while each class is paired with a human-annotated concept map and country-specific component descriptions and layout. 
SiFC supports direct zero-shot classification, training-free, and concept label-free CBM. 
It also supports testing out-of-region generalization.

\section{Method}
\label{sec:method}
\begin{figure}[h]
    \centering
    \includegraphics[width=\linewidth]
    {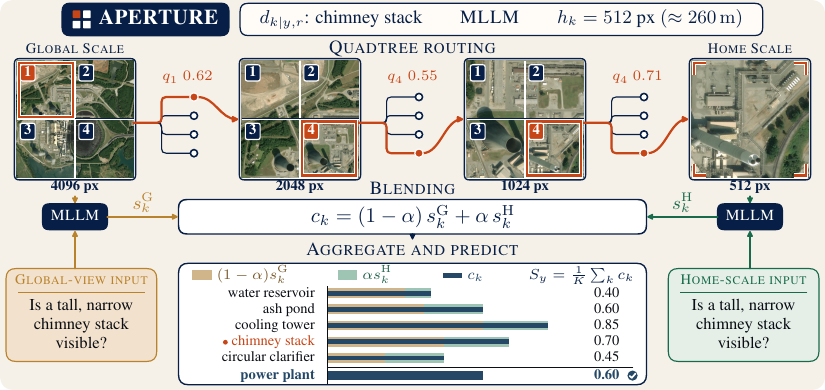}
    \caption{Overview of \methodname. For each class- and country-conditioned descriptor $d_{k\mid y,r}$, a frozen MLLM follows one quadtree branch to its home scale $h_k$ while pruning the remaining quadrants. Global and home-scale answer scores are blended into the concept score $c_k^{\mathrm{B}}$. The concept scores are then averaged into the class score $S_y$, and the highest-scoring class is predicted.}
    \label{fig:method-overview}
\end{figure}
We aim to build an interpretable-by-design recognition model for remote sensing categories. 
In \cref{ssec:background}, we first introduce variants of CBMs~\citep{pmlr-v119-koh20a}, as it forms the basis of our ante-hoc interpretable recognition method.
We use training-free (CBMs) as a starting point, and in \cref{ssec:mllm,ssec:prompts,ssec:concept-dependent-spatial-scale}, we introduce our contributions to improve training-free CBMs.

\subsection{Background}
\label{ssec:background}


\noindent\textbf{Concept Bottleneck Model:}
CBMs~\citep{pmlr-v119-koh20a} predict classes through an intermediate vector of interpretable concepts, $\hat{\mathbf{c}}=g(x)$, followed by a task predictor $\hat{y}=f(\hat{\mathbf{c}})$.
This intermediate representation enables interpretability and concept-level interventions.
However, supervised CBMs require costly concept-annotated triplets $\{x^{(i)},\mathbf{c}^{(i)},y^{(i)}\}_{i=1}^{N}$, especially in expert domains.

\noindent\textbf{CBMs with pre-trained VLM:}
Recent methods~\citep{oikarinen2023labelfree} use LLMs~\citep{brown2020language} or large concept dictionaries~\citep{oikarinen2023labelfree}, to construct a concept set $\mathcal{C}$ and to reduce annotation cost.
%
Pre-trained VLMs like CLIP models $g$, and image-class pairs are used $\{(x^{(i)},\mathbf{y}^{(i)}\}_{i=1}^{N}$ to learn $f$.
No expensive manual concept annotations are needed, but in many cases they learn incorrect correlations between classes and concepts that are not human-aligned.
In the following discussion we call these methods Concept Label-free CBMs (CLF-CBMs).

\noindent\textbf{Training-free CBMs:}
Training-free CBMs remove target-specific supervision entirely. They require no labeled target images, no concept annotations, and no parameter optimization on the target dataset. 
Like CLF-CBMs, LLMs construct a list of concepts; however, along with the list, the LLMs also provide a mapping between classes and specific concepts~\citep{menon2023visual}.
As a result, the image-concept function $g$, is replaced with a pre-trained VLM, and the concept-class function $f$ gets information from an LLM.
For each class $i$, the LLM selects a subset $\mathcal{C}_i \subseteq \mathcal{C}$ from the complete concept set.
Given an image's concept vector $\hat{\mathbf{c}}$, the class score is $\hat{y}_i = f_i(\hat{\mathbf{c}}) = \frac{1}{|\mathcal{C}_i|}\sum_{k\in\mathcal{C}_i}\hat{c}_k$, where $\hat{c}_k$ is the score for concept $k$.

We choose to build our method from TF-CBMs.
While TF-CBMs result in weaker-performing models compared to standard and CLF-CBMs, they remove expensive concept-label collection and target-specific training and are preferred in many cases, like ours, where data collection is expensive.
Moreover, in our results we observe that TF-CBMs also limit overfitting by avoiding learning spurious correlations. 
For our problem of satellite image classification, this is especially useful, as we show that TF-CBM models generalize much better to out-of-region images compared to LF-CBMs.
Furthermore, TF-CBM allows intervention and correction at the concept level without retraining.

\subsection{Problem Setup and Overview}
\label{ssec:overview}
Given a satellite image $x$ and country metadata $r$, we predict its facility class $y\in\mathcal{Y}$ without labeled target images, image-level concept annotations, or target-data training. 
We build on classification-by-description (CbD)~\citep{menon2023visual}, a TF-CBM. 
It represents each class using a set of concepts/descriptors, with the prepended class name, and uses CLIP as $g$.

However, applying CbD to satellite imagery presents two challenges. 
While CLIP and similar contrastive vision-language models understand global image concepts well, they struggle to understand part and local attributes~\citep{kang2025clip}.
This problem is further exacerbated when the concepts belong to expert domains like satellite images.
Another challenge is that industrial facilities span large areas and contain components at widely varying scales, so passing the full image to the vision encoder may obscure small but discriminative structures (e.g., a storage tank can be recognized at a coarse facility scale, whereas thin flare stacks require a finer local scale). 

These challenges motivate us to revise how concepts are recognized. 
To tackle the first challenge, we replace CLIP-like VLMs with general multimodal large-language models (MLLM).
We discuss the details of these modifications in \cref{ssec:mllm}.
Replacing contrastive VLMs with MLLMs also requires updating the concept definitions; we discuss them in \cref{ssec:prompts}.
To tackle the second challenge, we present a methodology to perform concept recognition at multiple scales in \cref{ssec:concept-dependent-spatial-scale}

\subsection{From CLIP to MLLM}
\label{ssec:mllm}
CbD scores each concept with a zero-shot CLIP-style encoder, but it does not work well on satellite images. 
CLIP~\citep{radford2021learning} itself reports that it is ``quite weak on satellite image classification (EuroSat~\citep{8736785} and RESISC45 ~\citep{cheng2017remote})''. 
This issue is further amplified when CLIP is used for recognizing localized concepts instead of global classes.
Furthermore, since remote-sensing VLMs~\citep{mall2024remote, liu2024remoteclip} are trained with similar methodology, they also fail to recognize concepts, even though they can recognize classes.

Since CbD relies on this alignment to score descriptors, it inherits the same limitation. 
We therefore replace CLIP with an MLLM.
We choose general-purpose MLLMs (e.g., Gemma~\citep{team2026gemma}), instead of remote sensing MLLMs, again, as remote sensing MLLMs are better at understanding classes in satellite images, but general-purpose MLLMs are much better at concepts.
To check for the presence of a concept (say, ``thin flare stacks''), instead of measuring the cosine similarity via CLIP, we ask an MLLM: 
\texttt{In the image is a thin flare stack present?}


\noindent\textbf{From binary answers to continuous concept scoring:}
\label{calibration}
However, when replacing a contrastive VLM with an MLLM, one challenge is that we do not get continuous scores and instead get binary answers.
CbD and CBM inputs require continuous concept scores~\citep{pmlr-v119-koh20a,espinosa2022concept,menon2023visual} and binarizing concept scores results in poorer performance, as it results in frequent ties and treats weak and strong evidence equally. 

One solution is to ask the model to report a confidence value with its answer (\texttt{e.g., ``yes, with 90\% confidence''}). 
However, we find that the verbalized confidence is unreliable because LLMs and VLMs are overconfident and poorly calibrated~\citep{xiong2024can}. 
Across 35,200 answers, Gemma-4-31B assigned 98.7\% of them a score of at least 0.9 and produced similar mean confidence for present and absent answers, 0.934 and 0.933, respectively. 
These scores provide less variance, and as a result, the model behaves like a binary decision-based CBM (Table~\ref{tab:headline-results}).

We instead present an alternative strategy to obtain continuous scores from MLLMs.
We compute the raw decoding logit scores when answering \texttt{yes} and \texttt{no}.
For each descriptor $d_k$, we construct the binary question
``\texttt{Is } $d_k$ \texttt{ visible in this satellite image? Answer yes or no.}''
We obtain the logits for \texttt{yes} ($z_1$) and \texttt{no} ($z_0$), which we convert into a continuous concept score using the temperature-scaled softmax
$p(\texttt{yes}\mid x)=\exp(z_1/\tau)\big/\left[\exp(z_1/\tau)+\exp(z_0/\tau)\right]$.
Compared with verbalized confidence, binary-answer probabilities provide a better-calibrated estimate of model uncertainty. 
Although decoding probabilities have previously been used to measure model uncertainty~\citep{kadavath2022language}, we are the first to use them in concept bottlenecks. 

\subsection{Changes to the concept Definitions}
\label{ssec:prompts}
Contrastive VLMs prefer shorter text queries (limited to 64 tokens for CLIP); so LLM-generated descriptors are typically intentionally kept shorter.
However, MLLMs support longer contexts and better grounding of sentences and paragraphs.
Therefore, we adapt our prompting as follows.
For each of these modifications, more details and exact prompts are present in appendix~\ref{app:concept-questioning}.

\noindent\textbf{From Attribute Phrases to Structured Descriptors.}
We observe that MLLMs respond more reliably to complete sentences with explicit visual cues (\cref{tab:headline-results}). 
We therefore expand each descriptor with additional details about \emph{appearance, spatial placement, and likely confusers.}
This structure provides both supporting and contrasting evidence, making concept verification less ambiguous. 


\noindent\textbf{From Explicit Class Names to Class- and Country-Conditioned Descriptors.}
The original CbD prepends class names to the descriptors, improving context and thus performance but hurting interpretability.
In our case, prepending labels not only limits interpretability, it also reduces performance.
Therefore, instead of prepending the class name, we prompt the LLM to modify the description conditioned on the class name.
Facilities of the same class can differ across countries; therefore, for generalization, we can also condition descriptions on country-specific information.

\noindent\textbf{Concept-Specific Home-Scale Assignment.}
Unlike internet images, objects in satellite imagery often have a well-defined physical scale. We exploit this \emph{home scale} to search for concepts at an appropriate scale. The descriptor-generation LLM assigns each descriptor a home scale $h_k\in\mathcal{S}$, defining the crop size that balances concept visibility and recognition context. 

\subsection{Concept-Dependent Spatial Scale.}
\label{ssec:concept-dependent-spatial-scale}
Satellite image concepts could cover different ground areas. 
Large concepts require a wide view to preserve their shape and layout. 
Small but important concepts require a local ``zoom-in'' for a local view to remain visible. 
This object scale variation is a known challenge in aerial imagery~\citep{8578516}, and prior work shows that remote-sensing representations benefit from explicit scale information~\citep{10377166, revankar2025scale}. 
We observe when a large $4\,\mathrm{km}^2$ tile is resized to the input resolution, small concepts are difficult to resolve. 
An LNG storage sphere (small concept) may occupy only two or three pixels, making it harder for the MLLM to recognize~\citep{khayatkhoei2025mllms,wang2025xlrs}.
Therefore, we do detection at multiple scales. 

One option is to zoom in to each concept's home scale, score every crop along the way, and keep the highest score at each level. 
We use crop widths in $\mathcal{S}=\{4096,2048,1024,512,256\}$ pixels. 
Starting from the full image, we stop zooming when we reach the concept's home scale $h_k\in\mathcal{S}$. 
Scoring every crop is costly, as crops increase exponentially with every zoom level. 
We therefore use a greedy \emph{quadtree search}; at each level, the MLLM selects the quadrant most likely to contain the concept. 
We zoom into that quadrant until we reach $h_k$. 
Each step halves the crop width (e.g., $4096\rightarrow2048\rightarrow1024$ pixels), so we search only one path through the quadtree.  
MLLM-based routing decides where to look, but we do not use its concept score at intermediate levels. 
At $h_k$, we ask whether the concept is visible in the selected crop and use $p(\texttt{yes})$. 
With $\ell_k=\log_2(4096/h_k)$ zoom levels, exhaustive search requires $O(4^{\ell_k})$ MLLM calls, while greedy routing requires $O(\ell_k+1)$; at $h_k=256$, this is $341$ versus $5$ calls.

The global view preserves facility layout, while the home-scale crop reveals small components. 
Therefore, we perform a linear blending of these global: $s_k^{\mathrm{G}}$ and  home-scale: $s_k^{\mathrm{H}}$ concept scores to get a final concept score: $c_k=(1-\alpha)s_k^{\mathrm{G}}+\alpha s_k^{\mathrm{H}}$.
Here, $\alpha\in[0,1]$ controls the home-scale contribution.
Like CbD, the class score is the average of individual \emph{blended} concept scores.

\section{Results}

\subsection{Implementation details and baselines}
\label{ssec:implementation}
We primarily evaluate \methodname~ on SiFC and the fMoW facility subset. 
Across tasks we report the macro-F1 (\%) score averaged over classes, both per-country and by pooling all images denoted by ``All''. 
Baselines with supervised training are trained on a leave-one-country-out setup. 

Global or home-scale views are resized to $224\times224$ images, matching CLIP ViT-L/14. 
Home-scale crops are extracted before resizing. 
Following prior binary QA work~\citep{giovannotti2024calibrated}, the temperature for question answering is set to $\tau=30$. 
Global and home-scale scores are blended with $\alpha=0.25$.
Our sensitivity analysis in the appendix shows that $\alpha$ values between $[0.2, 0.5]$, result in similar model performance.
Unless specified, Gemma-4-31B~\citep{team2026gemma}is our concept recognition backbone, but we also test other backbones in the ablation.

We compare with several training-free CBM baselines: zero-shot CLIP~\citep{radford2021learning}, CbD~\citep{menon2023visual}, and LaZSL~\citep{chen2025interpretable}, and supervised concept label-free CBMs: CF-CBM~\citep{panousis2024coarse}, LaBo~\citep{yang2023language}, and VLG-CBM~\citep{srivastava2024vlg}. 
We also compare two MLLM concept-scoring strategies: \emph{conf.}, which asks MLLM for a confidence score and \emph{prob.,} is our method using the next-token probability.
We also compare separate global and home-scale concept predictions and their blend (\methodname~) and also compare prompting strategies ranging from general to class-country-conditioned descriptors. 
\subsection{Discussion}
\label{ssec:discussion}

\providecommand{\ours}[1]{\textcolor{ourblue}{#1}}   
\begin{table}[h]
\centering
\caption{Macro-F1 (\%; $\uparrow$) on SiFC and the fMoW facility subset. The highest and second-highest scores in each column across all rows are \textbf{bold} and \underline{underlined}, respectively. All Gemma settings use the same frozen MLLM. TF denotes no task-specific training.}
\label{tab:headline-results}
\fontsize{7.5pt}{9pt}\selectfont
\setlength{\tabcolsep}{1.7pt}
\renewcommand{\arraystretch}{1.08}
\definecolor{ourtfink}{RGB}{18,78,59}
\definecolor{ourtfcanvas}{RGB}{234,245,238}
\renewcommand{\ours}[1]{\textcolor{ourtfink}{#1}}
\pgfmathsetlengthmacro{\ourtfrowheight}{\arraystretch*\ht\strutbox}
\pgfmathsetlengthmacro{\ourtfrowdepth}{\arraystretch*\dp\strutbox}
\begin{tikzpicture}[remember picture,overlay]
  \fill[ourtfcanvas]
    ([xshift=-1.5pt,yshift=\ourtfrowheight]pic cs:main-gemma-tf-start)
    rectangle ([yshift=-\ourtfrowdepth]pic cs:main-gemma-tf-end);
\end{tikzpicture}%
\begin{tabular*}{0.95\linewidth}{@{\extracolsep{\fill}}c@{\hspace{3pt}}l c l c@{\hspace{5pt}}rrr@{\hspace{4pt}}r@{\hspace{6pt}}rrr@{\hspace{4pt}}r@{}}
\toprule
\multirow{2}{*}{\textbf{Model}} & \multirow{2}{*}{\textbf{Setting}}
& \multirow{2}{*}{\textbf{TF}} & \multirow{2}{*}{\textbf{Descriptor}}
& \multirow{2}{*}{\textbf{Score}}
& \multicolumn{4}{c}{\textbf{SiFC}} & \multicolumn{4}{c}{\textbf{fMoW facility}} \\
\cmidrule(lr){6-9}\cmidrule(l){10-13}
& & & & & India & USA & China & \textbf{All} & USA & France & Russia & \textbf{All} \\
\midrule
\multirow{7}{*}{\rotatebox[origin=c]{90}{Baselines}} & CLIP zero-shot & \ding{51} & CND & cos. & 53.03 & 67.19 & 53.29 & 62.16 & 71.29 & 57.71 & 51.91 & 65.85 \\
 & CBD & \ding{51} & CBD & cos. & 59.37 & 66.60 & 57.73 & 66.42 & 70.57 & 54.11 & 51.02 & 62.74 \\
 & LaZSL & \ding{51} & AD & OT & 69.38 & 72.83 & 58.78 & 69.76 & \textbf{78.63} & 59.86 & 57.17 & 69.90 \\
\cmidrule(l){2-13}
 & CF-CBM (high) & \ding{55} & HCD & cos. & 40.25 & 25.67 & 48.60 & 39.35 & 12.60 & 25.89 & 24.11 & 23.18 \\
 & CF-CBM (low) & \ding{55} & HCD & cos. & 53.92 & 40.87 & 60.69 & 49.88 & 41.30 & 72.52 & 58.17 & 55.92 \\
 & LaBo & \ding{55} & LC & cos. & 66.70 & 62.97 & 53.39 & 60.44 & 64.75 & 57.50 & 65.21 & 65.11 \\
 & VLG-CBM & \ding{55} & LC & logit & \underline{71.91} & 64.04 & 66.88 & 68.21 & 69.40 & 68.26 & 62.28 & 71.62 \\
\midrule
 & Class name only & \ding{51} & CND & -- & 60.62 & 68.02 & 62.89 & 70.40 & 67.52 & 74.98 & 61.46 & 73.18 \\
\cmidrule(l){2-13}
 & \tikzmark{main-gemma-tf-start}\makebox[17pt][l]{\multirow{7.5}{*}[-0.5pt]{{\setlength{\fboxsep}{2pt}\colorbox{ourtfcanvas}{\rotatebox[origin=c]{90}{\makebox[68pt][c]{\ours{\textbf{{Our TF Setting}}}}}}}}}\ours{Home scale} & \ding{51} & HSR-CCCD & prob. & 41.74 & 66.10 & 45.56 & 55.09 & 57.76 & \underline{77.01} & 49.04 & 63.46 \\
\addlinespace[1pt]
 & \hspace*{17pt}\ours{Global Scale} & \ding{51} & RAG-CCCD & prob. & 70.68 & \underline{75.54} & 74.67 & 77.53 & 69.69 & 75.17 & \underline{70.07} & 74.40 \\
 &  & \ding{51} & CCCD & prob. & 69.36 & 74.81 & \underline{77.14} & \underline{78.62} & 72.63 & 75.41 & 67.16 & \underline{75.65} \\
 &  & \ding{51} & CCCD & conf. & 67.16 & 71.83 & 73.56 & 75.93 & 68.03 & 62.07 & \textbf{74.63} & 67.87 \\
 &  & \ding{51} & GCD & prob. & 60.72 & 71.74 & 64.06 & 70.78 & 72.99 & 72.01 & 68.06 & 73.90 \\
 &  & \ding{51} & GCD & conf. & 58.57 & 65.34 & 55.20 & 65.44 & 69.89 & 70.17 & 62.79 & 72.87 \\
\cmidrule(l{17pt}){2-13}
 & \hspace*{17pt}\ours{\textbf{\textsc{Aperture}}} & \ours{\ding{51}} & \ours{HSR-CCCD} & \ours{prob.} & \ours{\textbf{72.37}} & \ours{\textbf{75.65}} & \ours{\textbf{79.62}} & \ours{\textbf{80.19}} & \ours{\underline{73.09}} & \ours{\textbf{77.70}} & \ours{68.65} & \ours{\textbf{76.55}}\tikzmark{main-gemma-tf-end} \\
\cmidrule(l){2-13}
 & CLF-CBM & \ding{55} & CCCD & prob. & 58.46 & 66.34 & 68.95 & 68.15 & 71.99 & 70.03 & 69.87 & 72.90 \\
 &  & \ding{55} & CCCD & conf. & 70.74 & 68.62 & 73.71 & 74.36 & 61.65 & 74.65 & 61.80 & 69.82 \\
 &  & \ding{55} & GCD & prob. & 60.75 & 68.06 & 54.53 & 64.82 & 68.61 & 71.87 & 64.54 & 71.01 \\
\multirow{-12}{*}[4pt]{\rotatebox[origin=c]{90}{Gemma 4 31B}} &  & \ding{55} & GCD & conf. & 69.88 & 68.21 & 60.78 & 66.53 & 62.40 & 76.23 & 64.28 & 70.10 \\
\bottomrule
\end{tabular*}

\vspace{3pt}
\begin{minipage}{0.98\linewidth}
\scriptsize
\textit{Descriptors.} CND: the class name alone. CCCD: class- and country-conditioned concept descriptors, written by Gemini 3.1 Pro and scored on the human-reviewed concept map. RAG-CCCD: CCCD grounded in country-specific documents. GCD: 44 general concept descriptors shared by all classes and countries. HSR: home-scale routing, each concept read at its own scale. HCD: hierarchical concept descriptors (high: facility level, low: component level). LC: language-generated concepts. AD: attribute descriptions.
\textit{Scores.} prob.: probability of \emph{yes} from the yes/no token logits. conf.: confidence stated by the model. cos.: cosine similarity. OT: locally aligned optimal transport.
\textit{Settings.} Class name: zero-shot with the class name. Our training-free settings use the home-scale and global image, or their blend (\textsc{Aperture}). LF-CBM trains a concept bottleneck head on two countries and tests on the third.
\end{minipage}
\end{table}

\noindent\textbf{How well does our method perform?}
\label{ssec:model_perform}
We show the performance of our method and its variants in \cref{tab:headline-results}.
Aperture reaches 80.19\% macro-F1 on SiFC and 76.55\% on the fMoW facility subset (Table~\ref{tab:headline-results}). 
The performance is significantly better than any of the baselines.
On SiFC, \methodname~beats the best-performing baseline LaZSL by more than 10 pp, on fMoW VLG-CBM, by more than 3 pp.
From both datasets, we observe that combining local crops with the global view performs better than using either view alone.
For example, on SiFC All, home-scale and global-scale have an F1-score of $55.09\%$ and $78.62\%$, whereas combining the scale results in the best F1-score of $80.19\%$.

In most cases using decoding probabilities (\emph{prob.}) is better than MLLM-reported confidence (\emph{conf.}).
We also test the performance of different ways of creating descriptors.
Class- and country-conditioned descriptors (CCCD) improve the score by more than 8 pp, over global descriptors (GCD). 
We also further tried a RAG-based descriptor conditioning, where we use facility documentation from specific countries instead of LLM's prior knowledge, but this yields little improvement.



\noindent\textbf{Can our method generalize to other regions?}
\label{ssec:method_generalize}
Across countries, \methodname~ exceeds the best CLF-CBM in four of six comparisons (Table~\ref{tab:headline-results}). It performs better in India, the USA, and China on SiFC and France on fMoW-FC, while CLF-CBM is slightly stronger in the fMoW-FC USA and Russia. These results suggest that training-free CBMs generalize better to new regions, whereas supervised CLF-CBMs may overfit to their training countries.
\label{ssec:across_model}
\begin{figure}[h]
\centering
\begin{minipage}[t]{0.555\linewidth}
\vspace{0pt}
\centering
\captionsetup{font=scriptsize,labelfont=bf,justification=raggedright,singlelinecheck=false,position=top,skip=2pt}
\captionof{table}{CND vs.\ Aperture, macro-F1 (\%); $\Delta$ = Aperture $-$ CND (pp).}
\label{tab:cross-model}
\fontsize{7.2pt}{8.6pt}\selectfont
\setlength{\tabcolsep}{2.2pt}
\renewcommand{\arraystretch}{1.02}
\resizebox{\linewidth}{!}{%
\begin{tabular}{@{}cl*{4}{rr>{\color{deltared}}r}@{}}
\toprule
\multirow{2.8}{*}[2pt]{\rotatebox[origin=c]{90}{Dataset}}
& \multirow{2}{*}{Country}
& \multicolumn{3}{c}{Gemma 4}
& \multicolumn{3}{c}{GLM}
& \multicolumn{3}{c}{Qwen3.5}
& \multicolumn{3}{c}{Gemini 3.1} \\
\cmidrule(lr){3-5}
\cmidrule(lr){6-8}
\cmidrule(lr){9-11}
\cmidrule(l){12-14}
&
& CND & \methodname & $\Delta$
& CND & \methodname & $\Delta$
& CND & \methodname & $\Delta$
& CND & \methodname & $\Delta$ \\
\midrule
\multirow{4}{*}{\rotatebox[origin=c]{90}{SiFC}}
& India & 60.62 & 72.37 & $+11.75$ & 39.30 & 53.91 & $+14.61$ & 37.43 & 65.04 & $+27.61$ & 85.20 & 91.53 & $+6.33$ \\
& USA & 68.02 & 75.54 & $+7.52$ & 48.70 & 69.19 & $+20.49$ & 51.40 & 68.86 & $+17.46$ & 94.35 & 97.88 & $+3.53$ \\
& China & 62.89 & 79.62 & $+16.73$ & 48.71 & 60.41 & $+11.70$ & 44.07 & 57.48 & $+13.41$ & 87.20 & 91.12 & $+3.92$ \\
& All & 70.40 & 80.19 & $+9.79$ & 50.12 & 66.55 & $+16.43$ & 48.25 & 66.58 & $+18.33$ & 90.09 & 95.69 & $+5.60$ \\
\midrule
\multirow{4}{*}{\rotatebox[origin=c]{90}{fMoW}}
& USA & 67.52 & 72.32 & $+4.80$ & 66.11 & 68.07 & $+1.96$ & 62.47 & 67.72 & $+5.25$ & 80.27 & 76.97 & $-3.30$ \\
& France & 74.98 & 77.01 & $+2.03$ & 59.94 & 68.75 & $+8.81$ & 61.31 & 69.08 & $+7.77$ & 68.90 & 85.28 & $+16.38$ \\
& Russia & 61.46 & 68.65 & $+7.19$ & 56.96 & 62.73 & $+5.77$ & 67.82 & 62.53 & $-5.29$ & 83.89 & 91.36 & $+7.47$ \\
& All & 73.18 & 76.55 & $+3.37$ & 68.15 & 71.52 & $+3.37$ & 66.82 & 72.22 & $+5.40$ & 78.65 & 81.16 & $+2.51$ \\
\bottomrule
\end{tabular}%
}
\end{minipage}\hfill
\begin{minipage}[t]{0.425\linewidth}
\vspace{0pt}
\centering
\includegraphics[width=\linewidth]{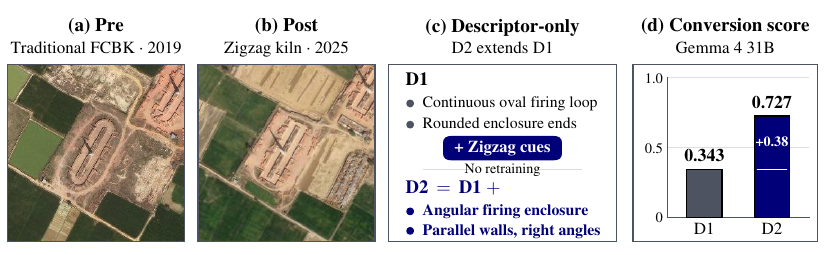}
\captionsetup{font=scriptsize,labelfont=bf,skip=2pt,justification=raggedright,singlelinecheck=false}
\caption{Descriptor-only adaptation to kiln conversion.
The kiln changes from FCBK to Zigzag.
Across 100 conversion sites, Zigzag-specific cues raise the
conversion score from 0.343 (D1) to 0.727 (D2),
with Gemma 4.}
\label{fig:temporal-kiln-conversion}
\end{minipage}
\end{figure}

\noindent\textbf{Does our method work across backbones?}
Combining global and home-scale views improves performance
across different MLLMs, not just Gemma-4-31B
(Tables~\ref{tab:cross-model} and~\ref{tab:full-results}).
Each model's blended result exceeds its own best global-only
concept setting wherever both were tested.
This holds for Gemma 4 31B, GLM-4.6V, Qwen3.5 122B-A10B,
and Gemini 3.1 Pro.
The complete results with these models are in the Appendix Table~\ref{tab:full-results}.

Although Gemini is the strongest backbone, evaluation with proprietary models is costly and less reproducible. 
We therefore report our main results using open-source MLLMs. 
Nevertheless, results show that our training-free framework also improves over zero-shot proprietary MLLMs.


\noindent\textbf{Does our method respond to visual interventions?}
\label{sec:intervention}
\methodname~ not only has a good performance, but it also allows better interpretability and intervention.
We study whether our method responds to visual interventions that remove class-defining parts. 
To test this, we select 90 images across six classes, we inpaint three parts one by one, and we measure the true-class score drop from the intact image. 
After the first removal, the mean drop is largest at the home scale (0.10), compared with global (0.04) and zero-shot scoring (0.03).
After three removals, concept-based scores drop more than zero-shot scores on average (0.43--0.49 versus 0.36; Table~\ref{tab:part-removal}).
These results show that \methodname~ is \emph{more sensitive to removed visual evidence}, with home-scale scoring responding strongest to part removal.
We also visualize individual images in the appendix ~\cref{fig:cumulative-coal-intervention,fig:cumulative-refinery-intervention,fig:cumulative-nuclear-intervention,fig:cumulative-lng-intervention,fig:cumulative-steel-intervention,fig:cumulative-sewage-intervention}.

\begin{table}[h]
\centering
\providecommand{\ours}[1]{#1}
\renewcommand{\ours}[1]{\textcolor{ourblue}{#1}}
\colorlet{prframe}{ourblue!80}
\definecolor{prbarzero}{RGB}{200,200,200}
\definecolor{prbarglobal}{RGB}{95,95,95}
\definecolor{prbarhome}{RGB}{20,60,170}
\definecolor{prbarblend}{RGB}{112,138,208}
\caption{\textbf{Cumulative part-removal interventions.}
Mean true-class score drop from the intact image over 15 images per class. S1--S3 cumulatively remove one, two, and three parts, respectively (listed below). The framed block averages six classes; bars end at S3, with cuts marking S1 and S2.\textbf{Bold}: largest drop per column; \ours{blue}: our settings.}
\label{tab:part-removal}
\fontsize{7pt}{8.5pt}\selectfont
\setlength{\tabcolsep}{0pt}
\renewcommand{\arraystretch}{1.2}
\begin{tabular}{@{}w{l}{34.237pt}@{\hspace{3.000pt}}w{r}{12.250pt}@{\hspace{2.400pt}}w{r}{12.250pt}@{\hspace{2.400pt}}w{r}{12.250pt}@{\hspace{4.000pt}}w{r}{17.640pt}@{\hspace{2.400pt}}w{r}{12.250pt}@{\hspace{2.400pt}}w{r}{12.250pt}@{\hspace{4.000pt}}w{r}{12.250pt}@{\hspace{2.400pt}}w{r}{12.250pt}@{\hspace{2.400pt}}w{r}{12.250pt}@{\hspace{4.000pt}}w{r}{12.250pt}@{\hspace{2.400pt}}w{r}{12.250pt}@{\hspace{2.400pt}}w{r}{12.250pt}@{\hspace{4.000pt}}w{r}{12.250pt}@{\hspace{2.400pt}}w{r}{12.250pt}@{\hspace{2.400pt}}w{r}{12.250pt}@{\hspace{4.000pt}}w{r}{17.640pt}@{\hspace{2.400pt}}w{r}{12.250pt}@{\hspace{2.400pt}}w{r}{18.040pt}@{\hspace{5.200pt}}w{r}{12.250pt}@{\hspace{2.400pt}}w{r}{12.250pt}@{\hspace{2.400pt}}w{r}{12.250pt}@{\hspace{2.600pt}}w{l}{22.178pt}@{\hspace{2.800pt}}}
\toprule
\multirow{2}{*}[-1.5pt]{\textbf{Method}} & \multicolumn{3}{c@{\hspace{4.000pt}}}{\makebox[41.550pt][c]{\textbf{Coal power}}} & \multicolumn{3}{c@{\hspace{4.000pt}}}{\makebox[46.940pt][c]{\textbf{Sewage}}} & \multicolumn{3}{c@{\hspace{4.000pt}}}{\makebox[41.550pt][c]{\textbf{Oil refinery}}} & \multicolumn{3}{c@{\hspace{4.000pt}}}{\makebox[41.550pt][c]{\textbf{LNG terminal}}} & \multicolumn{3}{c@{\hspace{4.000pt}}}{\makebox[41.550pt][c]{\textbf{Nuclear}}} & \multicolumn{3}{c@{\hspace{5.200pt}}}{\makebox[52.730pt][c]{\textbf{Iron \& steel}}} & \multicolumn{4}{c@{\hspace{2.800pt}}}{\tikzmark{prTop}\textbf{Mean}} \\
\cmidrule(l{0.000pt}r{4.000pt}){2-4}\cmidrule(l{0.000pt}r{4.000pt}){5-7}\cmidrule(l{0.000pt}r{4.000pt}){8-10}\cmidrule(l{-0.665pt}r{3.335pt}){11-13}\cmidrule(l{0.000pt}r{4.000pt}){14-16}\cmidrule(l{0.000pt}r{5.200pt}){17-19}\cmidrule(l{0pt}r{2.800pt}){20-23}
 & \makebox[12.25pt][c]{S1} & \makebox[12.25pt][c]{S2} & \makebox[12.25pt][c]{S3} & \makebox[12.25pt][c]{S1} & \makebox[12.25pt][c]{S2} & \makebox[12.25pt][c]{S3} & \makebox[12.25pt][c]{S1} & \makebox[12.25pt][c]{S2} & \makebox[12.25pt][c]{S3} & \makebox[12.25pt][c]{S1} & \makebox[12.25pt][c]{S2} & \makebox[12.25pt][c]{S3} & \makebox[12.25pt][c]{S1} & \makebox[12.25pt][c]{S2} & \makebox[12.25pt][c]{S3} & \makebox[12.25pt][c]{S1} & \makebox[12.25pt][c]{S2} & \makebox[12.25pt][c]{S3} & \makebox[12.25pt][c]{S1} & \makebox[12.25pt][c]{S2} & \makebox[12.25pt][c]{S3} &  \\
\cmidrule[\lightrulewidth](l{0pt}r{3.60pt}){1-19}\cmidrule[\lightrulewidth](l{-1.20pt}r{1.60pt}){20-23}
Zero-shot & 0.07 & 0.07 & 0.26 & \llap{$-$\kern-0.6pt}0.07 & 0.07 & 0.78 & 0.00 & 0.12 & 0.33 & \textbf{0.19} & 0.14 & 0.59 & 0.04 & 0.23 & 0.28 & \llap{$-$\kern-0.6pt}0.03 & 0.00 & \llap{$-$\kern-0.6pt}0.07 & 0.03 & 0.10 & 0.36 & \tikz[baseline=0pt]{\useasboundingbox(0,0)rectangle(22.178pt,4.37pt);\fill[prbarzero](0.000pt,0.37pt)rectangle(1.490pt,4.37pt);\fill[prbarzero](2.090pt,0.37pt)rectangle(4.603pt,4.37pt);\fill[prbarzero](5.203pt,0.37pt)rectangle(16.257pt,4.37pt);} \\
\cmidrule[\lightrulewidth](l{0pt}r{3.60pt}){1-19}\cmidrule[\lightrulewidth](l{-1.20pt}r{1.60pt}){20-23}
Global view & 0.05 & 0.22 & 0.26 & 0.04 & 0.14 & \textbf{0.84} & 0.01 & 0.22 & \textbf{0.70} & 0.02 & 0.20 & \textbf{0.73} & \textbf{0.09} & \textbf{0.25} & 0.35 & 0.02 & 0.04 & 0.07 & 0.04 & 0.18 & 0.49 & \tikz[baseline=0pt]{\useasboundingbox(0,0)rectangle(22.178pt,4.37pt);\fill[prbarglobal](0.000pt,0.37pt)rectangle(1.742pt,4.37pt);\fill[prbarglobal](2.342pt,0.37pt)rectangle(8.033pt,4.37pt);\fill[prbarglobal](8.633pt,0.37pt)rectangle(22.075pt,4.37pt);} \\
\ours{Home scale} & \ours{\textbf{0.14}} & \ours{\textbf{0.36}} & \ours{\textbf{0.45}} & \ours{\textbf{0.19}} & \ours{\textbf{0.35}} & \ours{0.72} & \ours{\textbf{0.01}} & \ours{\textbf{0.25}} & \ours{0.54} & \ours{0.13} & \ours{\textbf{0.31}} & \ours{0.54} & \ours{0.09} & \ours{0.21} & \ours{\textbf{0.47}} & \ours{\textbf{0.06}} & \ours{\textbf{0.15}} & \ours{\textbf{0.23}} & \ours{\textbf{0.10}} & \ours{\textbf{0.27}} & \ours{\textbf{0.49}} & \tikz[baseline=0pt]{\useasboundingbox(0,0)rectangle(22.178pt,4.37pt);\fill[prbarhome](0.000pt,0.37pt)rectangle(4.680pt,4.37pt);\fill[prbarhome](5.280pt,0.37pt)rectangle(12.233pt,4.37pt);\fill[prbarhome](12.833pt,0.37pt)rectangle(22.178pt,4.37pt);} \\
\ours{\textsc{Aperture}} & \ours{0.05} & \ours{0.21} & \ours{0.25} & \ours{0.06} & \ours{0.16} & \ours{0.75} & \ours{0.01} & \ours{0.18} & \ours{0.58} & \ours{0.03} & \ours{0.19} & \ours{0.62} & \ours{0.07} & \ours{0.22} & \ours{0.33} & \ours{0.02} & \ours{0.04} & \ours{0.06} & \tikzmark{prLeft}\ours{0.04} & \ours{0.17} & \ours{0.43} & \tikz[baseline=0pt]{\useasboundingbox(0,0)rectangle(22.178pt,4.37pt);\fill[prbarblend](0.000pt,0.37pt)rectangle(1.794pt,4.37pt);\fill[prbarblend](2.394pt,0.37pt)rectangle(7.521pt,4.37pt);\fill[prbarblend](8.121pt,0.37pt)rectangle(19.423pt,4.37pt);}\tikzmark{prRight} \\
\bottomrule
\end{tabular}%
\begin{tikzpicture}[remember picture,overlay]
  \coordinate (prT) at (pic cs:prTop);
  \coordinate (prL) at (pic cs:prLeft);
  \coordinate (prR) at (pic cs:prRight);
  \draw[prframe, line width=0.55pt, rounded corners=2.5pt]
    ([xshift=-2.4pt,yshift=7.9pt]prT -| prL)
    rectangle ([xshift=2.4pt,yshift=-2.9pt]prR);
\end{tikzpicture}%
\vspace{3pt}
\begin{minipage}{\linewidth}
\scriptsize
{\raggedright\textbf{Parts erased.} \textbf{Coal power:} S1\,$=$\,chimney; S2\,$=$\,S1\,$+$\,cooling tower; S3\,$=$\,S2\,$+$\,coal stockpile. \textbf{Sewage:} S1\,$=$\,sludge digester; S2\,$=$\,S1\,$+$\,aeration basin; S3\,$=$\,S2\,$+$\,circular clarifier. \textbf{Oil refinery:} S1\,$=$\,jetty; S2\,$=$\,S1\,$+$\,fixed-roof tanks; S3\,$=$\,S2\,$+$\,floating-roof tanks. \textbf{LNG terminal:} S1\,$=$\,regasification area; S2\,$=$\,S1\,$+$\,jetty; S3\,$=$\,S2\,$+$\,LNG tanks. \textbf{Nuclear:} S1\,$=$\,turbine hall; S2\,$=$\,S1\,$+$\,cooling tower; S3\,$=$\,S2\,$+$\,reactor dome. \textbf{Iron \& steel:} S1\,$=$\,blast furnace; S2\,$=$\,S1\,$+$\,raw material stockyard; S3\,$=$\,S2\,$+$\,long mill building.\par}\smallskip
\end{minipage}
\end{table}

\noindent\textbf{Can descriptors adapt to technical change?}
Along with the generated visual interventions, we also test a real-world use case.
We test whether descriptor updates can be used to track technological updates to facilities without retraining. 
Efforts to reduce brick-kiln pollution motivated conversion from FCBK-type to cleaner Zigzag technology~\citep{patel2025space}.
Across 100 conversion sites~\citep{mondal2026sentinelkilndb}, we compare original descriptors (D1) with descriptors extended by Zigzag-specific cues (D2).
The conversion score increases from 0.343 to 0.727 (Fig.~\ref{fig:temporal-kiln-conversion}), suggesting that textual interventions can improve recognition of such changes.

\noindent\textbf{Does our method work on the non-satellite CBM benchmark?}
We also evaluate \methodname~on CUB~\citep{wah2011caltech}, a standard internet-image CBM benchmark. On a 600-image, 20-class subset, our method outperforms standard CbD by 5 pp but remains below CLIP-based LaZSL (68.68\%; Table~\ref{tab:cub20-results}). We attribute this gap to LaZSL's random-crop strategy, which is better suited to datasets without a meaningful home scale.
\par


\subsection{Ablation}
\label{ssec:ablation}

\begin{table}[t]
\centering
\captionsetup{font=scriptsize,labelfont=bf,justification=raggedright,
  singlelinecheck=false,position=top,skip=2pt}
\setlength{\aboverulesep}{1pt}
\setlength{\belowrulesep}{1pt}
\begin{minipage}[t]{0.255\linewidth}
\vspace{0pt}
\caption{Training-free CUB-20: macro-F1 (\%) on 600 test images.}
\label{tab:cub20-results}
\fontsize{7pt}{8pt}\selectfont
\setlength{\tabcolsep}{0pt}
\renewcommand{\arraystretch}{1}
\begin{tabular*}{\linewidth}{@{\extracolsep{\fill}}llr@{}}
\toprule
Model & Setting & F1 \\
\midrule
\multirow{3}{*}{\shortstack[l]{CLIP\\L/14}}
 & Zero-shot & 57.69 \\
 & CBD & 59.80 \\
 & LaZSL & 68.68 \\
\midrule
\multirow{6}{*}{\shortstack[l]{Gemma\\4 31B}}
 & Class name & 34.62 \\
 & 4-crop CCD & 35.17 \\
 & 16-crop CCD & 35.17 \\
 & Global CCD & 62.46 \\
 & Global + 4 crops & 65.81 \\
 & Global + 16 crops & 64.24 \\
\bottomrule
\end{tabular*}
\end{minipage}\hfill
\begin{minipage}[t]{0.315\linewidth}
\vspace{0pt}
\caption{\textbf{Crop scale on SiFC.} Macro-F1 (\%); \textbf{All} is pooled.}
\label{tab:scale-ablation}
\fontsize{7pt}{8pt}\selectfont
\setlength{\tabcolsep}{0pt}
\renewcommand{\arraystretch}{1.3839}
\begin{tabular*}{\linewidth}{@{\extracolsep{\fill}}lrrr>{\bfseries}r@{}}
\toprule
\textbf{Crop/view} & \textbf{India} & \textbf{USA} & \textbf{China} & All \\
\midrule
Fixed 2048 & 45.3 & 62.5 & 50.3 & 54.9 \\
Fixed 1024 & 39.2 & 59.9 & 49.2 & 51.1 \\
Fixed 512 & 24.7 & 52.6 & 40.1 & 40.6 \\
\midrule
Assigned & 46.7 & 60.7 & 42.1 & 51.3 \\
Routed & 41.7 & 66.1 & 45.6 & 55.1 \\
\midrule
Global & 69.4 & 74.8 & 77.1 & 78.6 \\
\bottomrule
\end{tabular*}
\end{minipage}\hfill
\begin{minipage}[t]{0.385\linewidth}
\vspace{0pt}
\caption{\textbf{Training setups on SiFC.} Macro-F1 (\%); \textbf{All} is pooled.}
\label{tab:in-domain-cbm}
\fontsize{7pt}{8pt}\selectfont
\setlength{\tabcolsep}{0pt}
\renewcommand{\arraystretch}{1.6146}
\begin{tabular*}{\linewidth}{@{\extracolsep{\fill}}llrrr>{\bfseries}r@{}}
\toprule
\textbf{Setting} & \textbf{Desc.} & \textbf{India} & \textbf{USA} & \textbf{China} & All \\
\midrule
\multirow{2}{*}{In-domain}
 & CCCD & 74.99 & 88.04 & 82.56 & 84.53 \\
 & GCD & 71.14 & 87.04 & 82.95 & 83.98 \\
\midrule
\multirow{2}{*}{LF-CBM}
 & CCCD & 58.46 & 66.34 & 68.95 & 68.15 \\
 & GCD & 60.75 & 68.06 & 54.53 & 64.82 \\
\midrule
\textsc{Aperture} & HSR-CCCD & 72.37 & 75.65 & 79.62 & 80.19 \\
\bottomrule
\end{tabular*}
\end{minipage}
\end{table}

\noindent\textbf{What happens if we fix the home scale?}
We use Gemma 4 31B on 800 SiFC sites.
Fixed crops receive 113 questions each; assigned and routed views use scale-matched concepts.
Assigned maps 256-px concepts to 512-px crops and 4096-px concepts to the global view; the full 4096-px tile is resized to 224 px.
Fixed 2048 performs best (54.9\% macro-F1), close to routed home scale (55.1\%); assigned reaches 51.3\% and smaller fixed crops perform worse.
Global reaches 78.6\%, so context remains essential and routing adds little beyond the best fixed crop (Table~\ref{tab:scale-ablation}).

\noindent\textbf{How well does the model perform in-domain?}
In-domain heads use frozen concept probabilities and five-fold within-country out-of-fold evaluation, with inner 3-fold cross-validation selecting $C$.
LF-CBM trains on two countries and tests on the third; \textsc{Aperture} is training-free.
CCCD and GCD reach 84.53\% and 83.98\% macro-F1, versus 80.19\% for training-free \textsc{Aperture}.
These in-domain gains require labels from the evaluated country; LF-CBM instead tests cross-country (Table~\ref{tab:in-domain-cbm}).
Remaining ablations related to alternative search strategies, etc., are present in the Appendix.

\section{Conclusions}
We present \methodname~, a training-free concept bottleneck model for interpretable recognition in satellite images.
We show that both replacing contrastive VLMs with MLLMs and performing recognition at both global and characteristic home-scale results in significant improvement of interpretable training-free models.
We also introduce SiFC, a concept-centric recognition, and show that \methodname~ has much better out-of-region generalization than even supervised concept bottlenecks.
We also show that our method enables easy intervention and concept-level updates.
More broadly, our results suggest that effective concept-based recognition in expert domains requires reasoning about \emph{what} visual evidence defines a concept and \emph{where} to look for it.

\noindent \textbf{Limitations and future work} 
A limitation is that strong performance still requires blending home-scale predictions with global scores. 
Ideally, for component-level interpretability, home-scale evidence alone should suffice. 
Future work will explore strategies reducing reliance on global context.

\clearpage

\section*{Acknowledgments}
We thank the Anusandhan National Research Foundation (ANRF) for supporting this work through grant ANRF/ARG/2025/002304/ENS. We also acknowledge funding from Mohamed bin Zayed University of Artificial Intelligence (MBZUAI), Abu Dhabi, UAE. 

\subsection*{AI use statement}

In this work, we used AI tools to assist with grammar and spelling corrections, as well as sentence refinement. We have reviewed all the AI-assisted text, and we take responsibility for the final content of this work,
including the text, claims, or artifacts produced with the aid of generative AI.




\subsection*{Ethics statement}

Our work is intended to support academic research and environmental
monitoring by NGOs~\cite{esriNonprofitProgram}.
Our dataset release will contain no satellite images, tiles, or crops.
We will provide facility coordinates sourced from public government
records, source references, and a Python script for authorized imagery
retrieval.
Prior environmental studies have used Esri imagery for land-cover
annotation and permafrost-change validation~\citep{lesiv2025global,nitze2025darts}.
Users must obtain imagery directly from Esri and comply with applicable
ArcGIS and third-party terms, including download, use, and attribution
requirements~\citep{esri2025master,waterman2023wayback}.
Research or nonprofit status does not remove these obligations,
and our code grants no rights to the imagery



\subsection*{Reproducibility statement}

An anonymous downloadable code can be accessed through the link presented in the abstract. 
Additionally, information about the implementation and experimental setup is described in~\cref{ssec:implementation}. 
The details of processing the dataset will be released as easy-to-use scripts upon acceptance.



\clearpage
\bibliography{iclr2027_conference}
\bibliographystyle{iclr2027_conference}
\newpage
\appendix
\section{Appendix}
\begin{table}[h]
\centering
\caption{Macro F1 (\%) on SiFC (six classes, 800 images; India, USA, China) and on the fMoW subset (USA, France, Russia). All rows share one MLLM and are training-free (TF \ding{51}) except the CBM rows, which fit a linear head with leave-one-country-out folds (\ding{55}). Within each model block, \textbf{bold} is the best and \underline{underline} the second best per column. The row in blue is our full method.}
\label{tab:full-results}
\fontsize{7pt}{8.4pt}\selectfont
\setlength{\tabcolsep}{2.6pt}
\renewcommand{\arraystretch}{1.08}
\begin{tabular}{@{}l l l l c rrrr rrrr@{}}
\toprule
& & & & & \multicolumn{4}{c}{\textbf{SiFC (ours)}} & \multicolumn{4}{c}{\textbf{fMoW}} \\
\cmidrule(lr){6-9}\cmidrule(l){10-13}
\textbf{Model} & \textbf{Setting} & \textbf{Descriptor} & \textbf{Score} & \textbf{TF} &
\textbf{India} & \textbf{USA} & \textbf{China} & \textbf{All} &
\textbf{USA} & \textbf{France} & \textbf{Russia} & \textbf{All} \\
\midrule


\multirow{6}{*}{\textbf{Baseline}} & CLIP zero-shot & CND & cos. & \ding{51} & 53.03 & \textbf{67.19} & 53.29 & 62.16 & \textbf{71.29} & 57.71 & 51.91 & \underline{65.85} \\
 & CBD & CBD & cos. & \ding{51} & 59.37 & \underline{66.60} & 57.73 & \underline{66.42} & \underline{70.57} & 54.11 & 51.02 & 62.74 \\
 & LaZSL & AD & OT & \ding{51} & \underline{69.38} & \textbf{72.83} & 58.78 & \textbf{69.76} & \textbf{78.63} & 59.86 & 57.17 & \underline{69.90} \\
\cmidrule(l{0.4em}r{0.4em}){2-13}
 & CF-CBM (high) & HCD & cos. & \ding{55} & 40.25 & 25.67 & 48.60 & 39.35 & 12.60 & 25.89 & 24.11 & 23.18 \\
 & CF-CBM (low) & HCD & cos. & \ding{55} & 53.92 & 40.87 & \underline{60.69} & 49.88 & 41.30 & \textbf{72.52} & 58.17 & 55.92 \\
 & LaBo& LC & cos. & \ding{55} & \underline{66.70} & 62.97 & 53.39 & 60.44 & 64.75 & 57.50 & \textbf{65.21} & 65.11 \\
 & VLG-CBM& LC & logit & \ding{55} & \textbf{71.91} & 64.04 & \textbf{66.88} & \textbf{68.21} & 69.40 & \underline{68.26} & \underline{62.28} & \textbf{71.62} \\
\midrule


\multirow{12}{*}{\textbf{Gemma 4 31B}} & Class name only & CND & -- & \ding{51} & 60.62 & 68.02 & 62.89 & 70.40 & 67.52 & 74.98 & 61.46 & 73.18 \\
 & Home scale only & HSR-CCCD & prob. & \ding{51} & 41.74 & 66.10 & 45.56 & 55.09 & 57.76 & \textbf{77.70} & 49.04 & 63.46 \\
\cmidrule(l{0.4em}r{0.4em}){2-13}
 & Global concepts & RAG-CCCD & prob. & \ding{51} & 70.68 & \textbf{75.65} & 74.67 & 77.53 & 69.69 & 75.17 & \underline{70.07} & 74.40 \\
 &  & CCCD & prob. & \ding{51} & 69.36 & 74.81 & \underline{77.14} & \underline{78.62} & 72.63 & 75.41 & 67.16 & \underline{75.54} \\
 &  & CCCD & conf. & \ding{51} & 67.16 & 71.83 & 73.56 & 75.93 & 68.03 & 62.07 & \textbf{74.63} & 67.87 \\
 &  & GCD & prob. & \ding{51} & 60.72 & 71.74 & 64.06 & 70.78 & \textbf{73.05} & 72.01 & 68.06 & 73.90 \\
 &  & GCD & conf. & \ding{51} & 58.57 & 65.34 & 55.20 & 65.44 & 69.89 & 70.17 & 62.79 & 72.87 \\
\cmidrule(l{0.4em}r{0.4em}){2-13}
 & \textcolor{ourblue}{\textbf{Global + home (ours)}} & \textcolor{ourblue}{HSR-CCCD} & \textcolor{ourblue}{prob.} & \textcolor{ourblue}{\ding{51}} & \textcolor{ourblue}{\textbf{72.37}} & \textcolor{ourblue}{\underline{75.54}} & \textcolor{ourblue}{\textbf{79.62}} & \textcolor{ourblue}{\textbf{80.19}} & \textcolor{ourblue}{72.32} & \textcolor{ourblue}{\underline{77.01}} & \textcolor{ourblue}{68.65} & \textcolor{ourblue}{\textbf{76.55}} \\
\cmidrule(l{0.4em}r{0.4em}){2-13}
 & LF-CBM & CCCD & prob. & \ding{55} & 58.46 & 66.34 & 68.95 & 68.15 & \underline{72.99} & 70.03 & 69.87 & 72.90 \\
 &  & CCCD & conf. & \ding{55} & \underline{70.74} & 68.62 & 73.71 & 74.36 & 61.65 & 74.65 & 61.80 & 69.82 \\
 &  & GCD & prob. & \ding{55} & 60.75 & 68.06 & 54.53 & 64.82 & 68.61 & 71.87 & 64.54 & 71.01 \\
 &  & GCD & conf. & \ding{55} & 69.88 & 68.21 & 60.78 & 66.53 & 62.40 & 76.23 & 64.28 & 70.10 \\
\midrule
\multirow{11}{*}{\textbf{GLM-4.6V}} & Class name only & CND & -- & \ding{51} & 39.30 & 48.70 & 48.71 & 50.12 & 66.11 & 59.94 & 56.96 & \underline{68.15} \\
 & Home scale only & HSR-CCCD & prob. & \ding{51} & 30.62 & 54.22 & 33.20 & 40.97 & 62.59 & \underline{66.27} & 54.12 & 63.97 \\
\cmidrule(l{0.4em}r{0.4em}){2-13}
 & Global concepts & CCCD & prob. & \ding{51} & 52.16 & 63.63 & 60.89 & 65.02 & 65.50 & 60.04 & \underline{58.97} & 67.10 \\
 &  & CCCD & conf. & \ding{51} & 46.83 & 56.47 & 57.45 & 60.62 & 61.26 & 54.55 & 34.42 & 56.84 \\
 &  & GCD & prob. & \ding{51} & 41.32 & 52.80 & 47.66 & 53.61 & 62.48 & 61.42 & 52.76 & 65.42 \\
 &  & GCD & conf. & \ding{51} & 35.59 & 50.08 & 36.73 & 46.29 & 61.22 & 48.70 & 45.79 & 55.61 \\
\cmidrule(l{0.4em}r{0.4em}){2-13}
 & \textcolor{ourblue}{\textbf{Global + home (ours)}} & \textcolor{ourblue}{HSR-CCCD} & \textcolor{ourblue}{prob.} & \textcolor{ourblue}{\ding{51}} & \textcolor{ourblue}{\underline{53.91}} & \textcolor{ourblue}{\textbf{69.19}} & \textcolor{ourblue}{60.41} & \textcolor{ourblue}{\textbf{66.55}} & \textcolor{ourblue}{\textbf{68.07}} & \textcolor{ourblue}{\textbf{68.75}} & \textcolor{ourblue}{\textbf{62.73}} & \textcolor{ourblue}{\textbf{71.52}} \\
\cmidrule(l{0.4em}r{0.4em}){2-13}
 & LF-CBM & CCCD & prob. & \ding{55} & 52.66 & 57.99 & \underline{64.71} & 62.60 & 63.91 & 52.46 & 52.41 & 62.80 \\
 &  & CCCD & conf. & \ding{55} & 51.33 & 62.29 & 59.18 & 60.30 & \underline{66.24} & 57.89 & 56.84 & 64.08 \\
 &  & GCD & prob. & \ding{55} & \textbf{55.88} & \underline{67.03} & \textbf{65.61} & \underline{66.54} & 63.45 & 56.39 & 50.37 & 63.89 \\
 &  & GCD & conf. & \ding{55} & 47.88 & 62.71 & 55.69 & 59.64 & \underline{66.24} & 57.89 & 56.84 & 64.08 \\
\midrule
\multirow{11}{*}{\shortstack[l]{\textbf{Qwen3.5}\\\textbf{122B-A10B}}} & Class name only & CND & -- & \ding{51} & 37.43 & 51.40 & 44.07 & 48.25 & 62.47 & 61.31 & \textbf{67.82} & 66.82 \\
 & Home scale only & HSR-CCCD & prob. & \ding{51} & 40.65 & 55.11 & 42.48 & 48.52 & 54.52 & 64.06 & 52.95 & 58.99 \\
\cmidrule(l{0.4em}r{0.4em}){2-13}
 & Global concepts & CCCD & prob. & \ding{51} & 58.65 & 62.60 & 53.36 & 61.78 & 67.49 & 66.68 & 59.52 & \underline{70.69} \\
 &  & CCCD & conf. & \ding{51} & \underline{60.05} & 62.55 & 55.26 & 63.24 & 54.52 & 63.21 & 52.95 & 58.99 \\
 &  & GCD & prob. & \ding{51} & 49.07 & 57.25 & 41.05 & 53.11 & \textbf{70.60} & 61.54 & 61.28 & 69.70 \\
 &  & GCD & conf. & \ding{51} & 38.72 & 44.18 & 40.56 & 45.82 & 65.44 & 49.60 & 43.96 & 56.09 \\
\cmidrule(l{0.4em}r{0.4em}){2-13}
 & \textcolor{ourblue}{\textbf{Global + home (ours)}} & \textcolor{ourblue}{HSR-CCCD} & \textcolor{ourblue}{prob.} & \textcolor{ourblue}{\ding{51}} & \textcolor{ourblue}{\textbf{65.04}} & \textcolor{ourblue}{\textbf{68.86}} & \textcolor{ourblue}{\underline{57.48}} & \textcolor{ourblue}{\textbf{66.58}} & \textcolor{ourblue}{67.72} & \textcolor{ourblue}{\underline{69.08}} & \textcolor{ourblue}{62.53} & \textcolor{ourblue}{\textbf{72.22}} \\
\cmidrule(l{0.4em}r{0.4em}){2-13}
 & LF-CBM & CCCD & prob. & \ding{55} & 57.12 & \underline{68.61} & 51.11 & 61.77 & 69.16 & 59.07 & \underline{65.30} & 68.97 \\
 &  & CCCD & conf. & \ding{55} & 56.00 & 61.91 & \textbf{63.69} & \underline{64.06} & 68.40 & 56.84 & 58.80 & 66.53 \\
 &  & GCD & prob. & \ding{55} & 59.80 & 63.83 & 50.06 & 59.74 & 63.23 & \textbf{70.55} & 61.21 & 68.37 \\
 &  & GCD & conf. & \ding{55} & 47.02 & 57.45 & 48.85 & 54.63 & \underline{69.77} & 56.98 & 56.95 & 67.79 \\
\midrule
\multirow{4}{*}{\textbf{Gemini 3.1 Pro}}
 & Class name only & CND & -- & \ding{51}
 & 85.20 & 94.35 & 87.20 & 90.09
 & 80.27 & 68.90 & 83.89 & 78.65 \\
\cmidrule(l{0.4em}r{0.4em}){2-13}
 & Global scale & CCCD & conf. & \ding{51}
 & 86.78 & 90.01 & 91.12 & 92.53
 & 73.28 & 80.31 & 89.32 & 78.32 \\
 & Home scale only & HSR-CCCD & conf. & \ding{51}
 & 72.23 & 88.89 & 77.85 & 74.66
 & 70.09 & 76.09 & 77.26 & 72.39 \\
 \cmidrule(l{0.4em}r{0.4em}){2-13}
 & \textcolor{ourblue}{\textbf{Global + home (ours)}}
 & \textcolor{ourblue}{HSR-CCCD}
 & \textcolor{ourblue}{conf.}
 & \textcolor{ourblue}{\ding{51}}
 & 91.53 & 97.88 & 91.12 & 95.67
 & \textcolor{ourblue}{\textbf{76.97}}
 & \textcolor{ourblue}{\textbf{85.28}}
 & \textcolor{ourblue}{\textbf{91.36}}
 & \textcolor{ourblue}{\textbf{81.16}} \\
\bottomrule
\end{tabular}

\vspace{3pt}
\begin{minipage}{\textwidth}
\scriptsize
\textbf{Descriptors.} CND: the class name $n_y$ alone. CCCD: class- and country-conditioned concept descriptors $d_{k\mid y,r}$ for concept $k$, class $y$ and country $r$, written by Gemini 3.1 Pro (v4) and scored on the human-reviewed concept map. RAG-CCCD: CCCD grounded in country-specific documents. GCD: 44 general concept descriptors $d_k$ shared by all classes and countries. HSR: home-scale routing, where each concept is read at its own scale.
\textbf{Scores.} prob.: probability of \emph{yes} from the yes/no token logits. conf.: confidence stated by the model in its answer.
\textbf{Gemini 3.1 Pro} is a closed-source reference. \textbf{Settings.} Class name only: zero-shot with $n_y$. Home scale only: HSR without the global view. Global concepts: the full tile at 224 px. Global + home: both views. CBM (LOCO): a concept bottleneck head trained on two countries and tested on the third. HCD: hierarchical concept descriptors, with facility-level (high) and component-level (low) descriptions. LC: language-generated concepts expressed as short visual phrases. AD: attribute descriptions. OT: locally aligned optimal-transport score.

\end{minipage}
\end{table}

\begin{table}[h]
\centering
\caption{Proposed class and country distribution of the fMoW-FC facility subset.}
\label{tab:fmow-distribution}
\small
\setlength{\tabcolsep}{3pt}
\renewcommand{\arraystretch}{1.05}
\begin{tabular}{@{}lrrrr@{}}
\toprule
\textbf{Class} & \textbf{USA} & \textbf{France}
& \textbf{Russia} & \textbf{Total} \\
\midrule
Shipyard                 & 15 & 6  & 5  & 26  \\
Solar farm               & 24 & 20 & 6  & 50  \\
Storage tank             & 74 & 66 & 47 & 187 \\
Water treatment facility & 26 & 43 & 27 & 96  \\
Wind farm                & 22 & 14 & 5  & 41  \\
\midrule
\textbf{Total} & \textbf{161} & \textbf{149}
& \textbf{90} & \textbf{400} \\
\bottomrule
\end{tabular}
\end{table}

\begin{table}[t]
\centering
\caption{SiFC facility sites by class and country.}
\label{tab:sifc-distribution}
\small
\setlength{\tabcolsep}{3pt}
\renewcommand{\arraystretch}{1.05}
\begin{tabular}{@{}lrrrr@{}}
\toprule
\textbf{Facility class} & \textbf{India} & \textbf{USA}
& \textbf{China} & \textbf{Total} \\
\midrule
Iron and steel plant   & 48  & 28 & 104 & 180 \\
LNG terminal           & 8   & 13 & 28  & 49  \\
Nuclear power plant    & 7   & 52 & 15  & 74  \\
Oil refinery           & 24  & 70 & 65  & 159 \\
Power plant            & 101 & 42 & 37  & 180 \\
Sewage treatment plant & 22  & 90 & 46  & 158 \\
\midrule
\textbf{Total}          & \textbf{210} & \textbf{295}
& \textbf{295} & \textbf{800} \\
\bottomrule
\end{tabular}
\end{table}
\begin{table}[h]
\centering
\caption{Government references for SiFC facility identity and locality.
individual verification.}
\label{tab:sifc-official-sources}
\footnotesize
\setlength{\tabcolsep}{3pt}
\renewcommand{\arraystretch}{1.1}
\begin{tabular}{@{}lccc@{}}
\toprule
\textbf{Class} & \textbf{India} & \textbf{USA} & \textbf{China} \\
\midrule
Iron and steel
& \href{https://www.pib.gov.in/newsite/PrintRelease.aspx?relid=77494}{PIB}
& \href{https://www.epa.gov/ghgreporting/ghgrp-2022-metals}{EPA}
& \href{https://www.mee.gov.cn/gkml/sthjbgw/spwj1/201406/t20140606_276598.htm}{MEE} \\
LNG terminals
& \href{https://pngrb.gov.in/pdf/confluence/SESSION-1-AUTHORISATION.pdf}{PNGRB}
& \href{https://www.ferc.gov/natural-gas/lng}{FERC}
& \href{https://www.mee.gov.cn/gkml/sthjbgw/spwj1/201404/t20140404_270154.htm}{MEE} \\
Nuclear plants
& \href{https://www.npcil.nic.in/WriteReadData/userfiles/file/Corporate_Profile_English_2025.pdf}{NPCIL}
& \href{https://www.eia.gov/electricity/data/eia860/index.php}{EIA-860}
& \href{https://nnsa.mee.gov.cn/ztzl/jgdxsjk/hdc/yxhdjz/}{NNSA} \\
Oil refineries
& \href{https://ppac.gov.in/infrastructure/installed-refinery-capacity}{PPAC}
& \href{https://www.eia.gov/petroleum/refinerycapacity/index.php}{EIA-820}
& \href{https://www.mee.gov.cn/xxgk2018/xxgk/xxgk11/201906/t20190605_705757.html}{MEE} \\
Power plants
& \href{https://cea.nic.in/wp-content/uploads/pdm/2023/01/List_of_Power_Stations_31.03.2022.pdf}{CEA}
& \href{https://www.eia.gov/electricity/data/eia860/index.php}{EIA-860}
& \href{https://www.mee.gov.cn/gkml/sthjbgw/spwj1/201401/t20140115_266448.htm}{MEE} \\
Sewage treatment
& \href{https://cpcb.nic.in/openpdffile.php?id=UmVwb3J0RmlsZXMvMTIyOF8xNjE1MTk2MzIyX21lZGlhcGhvdG85NTY0LnBkZg==}{CPCB}
& \href{https://echo.epa.gov/tools/data-downloads/icis-npdes-download-summary}{EPA ECHO}
& \href{https://www.mee.gov.cn/ywdt/gs/wqgs_1/201901/W020190115497466861869.pdf}{MEE} \\
\bottomrule
\end{tabular}
\end{table}
\subsection{Changes to Concept Questioning}
\label{app:concept-questioning}

\label{app:Structured_descriptors}
\begingroup
\definecolor{apromptink}{RGB}{18,48,72}
\definecolor{apromptbackground}{RGB}{248,250,252}
\setlength{\fboxsep}{7pt}
\setlength{\fboxrule}{0.4pt}

\newcommand{\apromptslot}[1]{%
  \textcolor{apromptink}{\texttt{\{#1\}}}%
}

\newcommand{\apromptbox}[2]{%
  \par\vspace{5pt}\noindent
  \fcolorbox{apromptink}{apromptbackground}{%
    \begin{minipage}{\dimexpr\linewidth-2\fboxsep-2\fboxrule\relax}
      \small\raggedright
      \setlength{\parindent}{0pt}
      \setlength{\parskip}{4pt}
      {\color{apromptink}\bfseries #1\par}
      {\color{apromptink}\hrule height 0.4pt}
      \vspace{3pt}
      #2
    \end{minipage}%
  }\par
}


\noindent\textbf{From Attribute Phrases to Structured Descriptors.}

We observe that instruction-following MLLMs respond more reliably to complete sentences with explicit visual cues~\ref{tab:headline-results}. 
We therefore expand each descriptor into a presence with \emph{appearance, spatial placement, and likely confusers.} e.g., instead of using only ``\texttt{cooling tower}'', we ask, ``\texttt{Is a large hyperbolic concrete shell with a thick rim and dark open center visible near the turbine hall? Do not confuse it with a storage dome, which has a solid roof and casts a shorter shadow.}''. 
This structure provides both supporting and contrasting evidence, making concept verification less ambiguous. 
The goal is not merely utilizing larger context windows but providing a clearer description of what the model should identify and exclude.

Each descriptor combines visual appearance with cues that distinguish
similar objects. The generation template below is reconstructed from
the saved descriptor structure; the example retains the saved question
wording with its scale placeholder filled.

\apromptbox{A. Descriptor-generation template (reconstructed)}{%
\textbf{Input concept:} \apromptslot{concept}

Expand this concept into a structured description for recognition
in overhead satellite imagery. Focus on the object itself without
naming the facility class.

\noindent\textbf{From Explicit Class Names to Class-Conditioned Descriptors}
The original CBD implementation prepends class names to the descriptors.
Prepending class names provides better context to a descriptor. 
For example, a ``leg of a table'' semantically is a very different concepts from ``leg of a lion''.
This results in improved performance but affects interpretability, as the final classification is not just a function of the concept bottleneck, and some indirect signal is coming from the class name encoded. Therefore, rather than prepending the class name, we  condition the concept descriptors on the class name.
Building on the previously derived structured descriptions, we condition each concept on the facility class, because the same component can appear differently across classes. 
Prepending labels not only limits interpretability, but in our experiments, prepending labels also reduces performance.
We construct $d_{k\mid y}$, a visual description of concept $k$ conditioned on class $y$, while omitting the class name from the verification question. A cooling-tower descriptor can be \texttt{``a cooling tower beside boiler buildings and coal stockpiles''} for power plants or \texttt{``a cooling tower beside a domed building"} for nuclear power plants.
This captures class-specific appearance while keeping verification focused on visible evidence.

\textbf{Appearance:} Describe its visible shape, approximate size,
colour, surface features, and shadow.

\textbf{Confusers:} Identify similar-looking objects and the visible
cues that distinguish them.

\textbf{Verification question:} Assemble the fields as follows:

This is an aerial satellite image, \apromptslot{scale\_text}.
Is a/an \apromptslot{concept} visible in it?
\apromptslot{appearance}
Do not confuse with: \apromptslot{confusers}
Answer with exactly one word: yes or no.
}

\apromptbox{B. Worked example: Chimney stack}{%
\textbf{Input concept:} Chimney stack.\quad
\textbf{Scale text:} about 2 km across.

\textbf{Verification question:}
This is an aerial satellite image, about 2 km across.
Is a chimney stack visible in it?
Appearing as a small 2 to 10 meter circular dot in gray, white,
or red-and-white stripes, a chimney stack is most easily
identified by its exceptionally long, needle-like shadow cast
across the ground.
Do not confuse with: It is most often mistaken for silos, water
towers, or flare stacks, but can be distinguished by its much
longer, narrower shadow and the occasional presence of a visible
smoke or steam plume emitting from the open top.
Answer with exactly one word: yes or no.
}

\label{app:Class-conditioned_descriptors}
\par\medskip
\noindent\textbf{From Explicit Class Names to Class-Conditioned Descriptors.}
The class guides the description of appearance, confusers, and
placement. The adapted verification question below uses these
visual cues without explicitly naming the class.

\apromptbox{C. Class-conditioned generation template (condensed)}{%
Write visual descriptors for overhead satellite imagery at
approximately 0.5 m per pixel.

\textbf{Concept:} \apromptslot{concept}

\textbf{Facility class:} \apromptslot{class}

\textbf{Generic description:}
Appearance: \apromptslot{appearance};
confusers: \apromptslot{confusers};
placement: \apromptslot{placement}.

Rewrite these fields for how the concept appears in this facility
class. Preserve valid details and specialize its size, count,
form, and surrounding structures where appropriate.

\textbf{Return only a JSON object with three fields:}

\texttt{appearance}: Visible shape, size, colour, shadow,
and typical count.

\texttt{confusers}: Similar-looking objects and the visible
details that distinguish them.

\texttt{placement}: Nearby structures and visible connections.

\textbf{Constraints:} Describe the class generally, not one site.
Retain generic wording when no class-specific difference exists.
Do not include the facility class name in the returned fields.
}

\apromptbox{D. Worked example: Chimney stack (adapted question)}{%
\textbf{Generation inputs:}
Concept: chimney stack;
class: coal or thermal power plant.

\textbf{Class-name-free verification question:}
This is an aerial satellite image, about 2 km across.
Is a chimney stack visible in it?
Appearing as one to four bright circular dots 15 to 30 m across
at the base, these structures are most easily identified by their
very long, needle-thin shadows of 100 to 300 m lying across the
site and the frequent presence of a visible plume.
Do not confuse with: Cooling towers also emit plumes but are
much wider with shorter, thicker shadows, while transmission
towers cast shorter, lattice-patterned shadows and stand in
linear networks.
Placement: Positioned immediately beside the main boiler
building or emissions control equipment, it is connected to
them by large, visible flue gas ducts.
Answer with exactly one word: yes or no.
}

\apromptbox{F. Worked example: The same concept across three countries}{%
\textbf{Shared generation inputs:}
Concept: chimney stack;
class: coal or thermal power plant.
Only the country input changes.

\textbf{Shared verification format:}
This is an aerial satellite image, about 260 m across.
Is a chimney stack visible in it?
\apromptslot{appearance}
Do not confuse with: \apromptslot{confusers}
Answer with exactly one word: yes or no.

\textbf{Country-conditioned appearance excerpts:}

\textbf{India.}
Appearing as \textbf{one to six} bright, \textbf{concrete-grey}
circular dots 15 to 30 m across at the base, these structures
are identifiable by their extremely long, needle-thin shadows
of \textbf{150 to 300 m} and the frequent presence of a
visible plume.

\textbf{China.}
Appearing as \textbf{one to four} bright \textbf{white or grey}
circular dots 15 to 30 m across, these structures are most
easily identified by their extremely long, needle-thin shadows
of \textbf{150 to over 300 m} lying across the site and the
frequent presence of a \textbf{dense white plume}.

\textbf{USA.}
Appearing as \textbf{one to three} bright, \textbf{light grey}
circular dots 15 to 30 m across at the base, these structures
are most easily identified by their very long, needle-thin
shadows of \textbf{150 to over 300 m} lying across the site
and the frequent presence of a visible plume
\textbf{if the unit remains active}.

\textit{Reading guide:}
Country labels identify generation inputs and are not included
in the object-verification question. Bold highlights differences
in the saved LLM-generated descriptions, not measured
country-level statistics. Only the appearance field is shown
for comparison.
}

\par\medskip
\noindent\textbf{Concept-Specific Home-Scale Assignment.}
The home scale specifies the source-image crop width used to
verify a concept. The template below is reconstructed; the
example reports a saved assignment.

\apromptbox{G. Home-scale assignment template (reconstructed)}{%
\textbf{Concept:} \apromptslot{concept}

\textbf{Descriptor:}
Appearance: \apromptslot{appearance};
confusers: \apromptslot{confusers};
placement: \apromptslot{placement}.

\textbf{Image setup:}
The source image is $4096\times4096$ pixels at approximately
$0.5$ m per pixel. Each selected crop is resized to
$224\times224$ pixels before verification.

\textbf{Task:}
Assign a crop width that balances visible detail with the
surrounding context needed to recognize this concept.

Choose from $4096$, $2048$, $1024$, $512$, or $256$ source pixels.
Use wider crops for large structures or spatial arrangements,
and tighter crops for small components. Consider the visual
cues needed to distinguish the concept from similar objects.

\textbf{Return only:}
A JSON object with the field \texttt{home\_scale\_px}
containing one of the allowed widths.
}

\apromptbox{H. Worked example: Chimney stack}{%
\textbf{Descriptor entry:}
Concept: chimney stack;
class: coal or thermal power plant;
country: India.

\textbf{Descriptor cues (condensed):}
Small, concrete-grey circular bases with long, needle-thin
shadows, positioned beside boiler buildings and connected
by visible flue gas ducts.

\textbf{Saved assignment:}
\texttt{\{"home\_scale\_px": 512\}}

\textbf{Use during inference:}
Greedy quadtree routing stops at a $512\times512$-pixel crop
from the original image. This crop is resized to
$224\times224$ pixels for concept verification.

\textit{Note:}
The home scale is an assigned setting, not an experimentally
established optimum.
}

\par\medskip
\noindent\textbf{Final Inference Prompts for \textsc{Aperture}.}
Each descriptor provides questions for global verification,
quadrant routing, and home-scale verification. The templates
below preserve the saved prompt structure; placeholders are
filled from the country-selected descriptor entry.

\apromptbox{I. Global verification}{%
This is an aerial satellite image, about 2 km across.
Is a/an \apromptslot{concept} of the kind found at
a/an \apromptslot{class\_name} visible in it?
\apromptslot{appearance}
Do not confuse with: \apromptslot{confusers}
Placement: \apromptslot{placement}
Answer with exactly one word: yes or no.
}

\apromptbox{J. Quadrant routing}{%
This aerial satellite image covers \apromptslot{scale\_text}
and is divided into four equal quadrants:
1 upper-left, 2 upper-right, 3 lower-left, and 4 lower-right.
Assume the target is somewhere in this crop.
Which ONE quadrant contains the clearest visual evidence
consistent with the concept \apromptslot{concept}
associated with \apromptslot{class\_name}?

Target appearance: \apromptslot{appearance}

Expected placement: \apromptslot{placement}

Do not confuse with: \apromptslot{confusers}

This is only a zoom-location decision, not a final yes-or-no
presence decision. If the evidence is weak, still choose the
quadrant containing the strongest candidate evidence.
Answer with exactly one digit: 1, 2, 3, or 4.
}

\apromptbox{K. Home-scale verification: object-level concepts}{%
This is an aerial satellite image, \apromptslot{scale\_text}.
Is a/an \apromptslot{concept} visible in it?
\apromptslot{appearance}
Do not confuse with: \apromptslot{confusers}
Answer with exactly one word: yes or no.
}

Routing uses a numbered grid; home-scale verification uses
the clean crop. The scale text follows the current crop width:
4096, 2048, 1024, 512, and 256 pixels correspond to
``about 2 km across'', ``about 1 km across'',
``about 500 m across'', ``about 260 m across'', and
``about 130 m across'', respectively.
For tile-level concepts, the saved home question retains the
global question structure with the appropriate scale text.
A home scale of 4096 reuses the global evidence.

\apromptbox{L. Saved example: Chimney stack in India}{%
\textbf{Descriptor metadata:}
Concept: chimney stack;
class: coal or thermal power plant;
country: India;
home scale: 512 pixels.

\textbf{Routing scales:}
$4096 \rightarrow 2048 \rightarrow 1024 \rightarrow 512$.
The routing question is asked at the first three scales,
followed by verification at 512 pixels.

\textbf{Saved home-scale question:}

This is an aerial satellite image, about 260 m across.
Is a chimney stack visible in it?
Appearing as one to six bright, concrete-grey circular dots
15 to 30 m across at the base, these structures are identifiable
by their extremely long, needle-thin shadows of 150 to 300 m
and the frequent presence of a visible plume.
Do not confuse with: Cooling towers also emit plumes but
are much wider with massive dark central openings and shorter,
thicker shadows, while transmission towers cast
lattice-patterned shadows and stand in linear networks.
Answer with exactly one word: yes or no.
}

\noindent\textbf{What if we use a different search strategy?}

\begin{table}[t]
\centering
\caption{\textbf{Search width on SiFC.}
Macro-F1 (\%); \textbf{All}: pooled.}
\label{tab:beam-routing-compact}
\scriptsize
\setlength{\tabcolsep}{2pt}
\begin{tabular}{lrrrrrrrr}
\toprule
 & \multicolumn{2}{c}{\textbf{Top-1}}
 & \multicolumn{2}{c}{\textbf{Beam-2}}
 & \multicolumn{2}{c}{\textbf{Beam-3}}
 & \multicolumn{2}{c}{\textbf{Beam-4}} \\
\cmidrule(lr){2-3}
\cmidrule(lr){4-5}
\cmidrule(lr){6-7}
\cmidrule(lr){8-9}
\textbf{Region}
 & Home & Blend
 & Home & Blend
 & Home & Blend
 & Home & Blend \\
\midrule
India & 41.74 & 72.37 & 49.13 & 71.26 & 53.04 & 71.55 & 52.11 & 70.41 \\
USA   & 66.10 & 75.54 & 73.13 & 77.91 & 76.40 & 78.27 & 74.00 & 77.94 \\
China & 45.56 & 79.62 & 53.17 & 80.48 & 53.31 & 79.01 & 53.84 & 79.67 \\
\midrule
\textbf{All}
 & \textbf{55.09} & \textbf{80.19}
 & \textbf{61.58} & \textcolor{blue}{\textbf{80.42}}
 & \textbf{64.13} & \textbf{80.11}
 & \textbf{63.10} & \textbf{80.12} \\
\bottomrule
\end{tabular}
\end{table}

We test wider search using frozen Gemma 4 31B, keeping the descriptors,
home scales, and SiFC images unchanged.
We expand greedy top-1 quadtree routing to beams of 2, 3, and 4
(Table~\ref{tab:beam-routing-compact}).
Home denotes home-only scoring; Blend combines global and home views.
Blend settings were selected using evaluation labels.
Home-only macro-F1 rises from 55.09\% to 64.13\% with beam 3.
The global--home blend changes little: beam 2 reaches 80.42\%,
only 0.23 percentage points above greedy (80.19\%),
while beams 3 and 4 score lower.
Wider beams add crop evaluations with little blend gain,
so we retain greedy top-1 routing.

\input{figures/cumulative_interventions_all_float}
\end{document}